%% file: main.tex
\documentclass[10pt,twocolumn,letterpaper]{article}

\usepackage[pagenumbers]{cvpr} 

\input{preamble}

\usepackage{booktabs}
\usepackage{siunitx}
\definecolor{cvprblue}{rgb}{0.21,0.49,0.74}
\usepackage[pagebackref,breaklinks,colorlinks,citecolor=cvprblue]{hyperref}

\def\paperID{362} 
\def\confName{3DV\xspace}
\def\confYear{2026\xspace}

\title{RL-AD-Net: Reinforcement Learning Guided Adaptive Displacement in Latent Space for Refined Point Cloud Completion}

\author{Bhanu Pratap Paregi\\
GeoAI4Cities Lab,\\
IISER Bhopal,India\\
{\tt\small bhanu24@iiserb.ac.in}
\and
Vaibhav Kumar\\
GeoAI4Cities Lab,\\
IISER Bhopal,India\\
{\tt\small vaibhav@iiserb.ac.in}
}

\begin{document}
\maketitle

\begin{abstract}
Recent point cloud completion models, including transformer-based, denoising-based, and other state-of-the-art approaches, generate globally plausible shapes from partial inputs but often leave local geometric inconsistencies. We propose \textbf{RL-AD-Net}, a reinforcement learning (RL) refinement framework that operates in the latent space of a pretrained point autoencoder. The autoencoder encodes completions into compact global feature vectors (GFVs), which are selectively adjusted by an RL agent to improve geometric fidelity. To ensure robustness, a lightweight non-parametric PointNN selector evaluates the geometric consistency of both the original completion and the RL-refined output, retaining the better reconstruction. When ground truth is available, both Chamfer Distance and geometric consistency metrics guide refinement. Training is performed separately per category, since the unsupervised and dynamic nature of RL makes convergence across highly diverse categories challenging. Nevertheless, the framework can be extended to multi-category refinement in future work. Experiments on ShapeNetCore-2048 demonstrate that while baseline completion networks perform reasonable under their training-style cropping, they struggle in random cropping scenarios. In contrast, \textbf{RL-AD-Net consistently delivers improvements} across both settings, highlighting the effectiveness of RL-guided ensemble refinement. The approach is lightweight, modular, and model-agnostic, making it applicable to a wide range of completion networks without requiring retraining.
\end{abstract}

\section{Introduction}

Point cloud completion seeks to reconstruct the full 3D geometry of an object from partial observations a fundamental capability for applications such as robotics, autonomous driving, AR/VR content creation, and digital heritage preservation. Real-world scan data often suffer from occlusions, sensor noise, and limited viewpoints; thus, accurate shape completion is critical to enable reliable downstream perception and rendering tasks.

Transformer-based architectures, for example AdaPoinTr~\cite{yu2022adapointr}, have achieved state-of-the-art performance by capturing long-range dependencies through set-to-set modeling and geometry-aware attention. These methods produce coherent global reconstructions, yet they still frequently exhibit residual artifacts, smoothed-out features, or missing thin structures issues that can impair utility in precise tasks such as grasping or collision inference.

A practical and lightweight solution to address these imperfections is post-hoc refinement enhancing completion outputs without retraining or modifying the underlying network. Prior approaches often resort to deterministic filtering, regression-based residual corrections, or optimization-based smoothing, but they lack adaptability to individual instances or the ability to leverage learned geometric priors.

Inspired by earlier attempts to apply reinforcement learning (RL) to completion tasks such as RL-GAN-Net~\cite{sarmad2019rl}, which uses an RL agent to control a GAN in latent space we propose a \textit{Reinforcement Learning Guided Adaptive Displacement in Latent
Space for Refined Point Cloud Completion} (\textbf{RL-AD-Net}), a latent-space RL refinement framework specifically designed for point cloud completion. Instead of coupling RL with a generative adversarial model, we operate in the embedding space of a lightweight, category-specific point cloud autoencoder (AE). Our method proceeds as follows: given a baseline completion, the completed point cloud is encoded into a compact global feature vector (GFV). A dedicated RL agent, trained per category, predicts a latent adjustment vector. The adjusted GFV is decoded via a frozen AE decoder to yield a refined point cloud. This strategy allows the agent to perform data-driven, instance-specific corrections that preserve overall structure while resolving local errors.

We adopt a category-wise design, training one AE and one RL agent per object class. We found empirically that training a joint AE/RL agent across multiple categories led to reduced refinement performance likely due to conflicting geometric distributions across classes. Moreover, many real-world applications (e.g., warehouse robotics, autonomous driving) focus on a limited set of relevant object categories rather than covering all 55 classes of ShapeNet. Thus, our approach is both empirically justified and practically efficient.

Although we demonstrate RL-AD-Net using AdaPoinTr and ShapeNetCore-2048, our method is fundamentally model-agnostic: it can be applied to outputs from any completion model.

\textbf{Our key Contributions}:
\begin{itemize}
  \item We introduce a novel \emph{category-wise latent-space RL refinement} method that enhances point cloud completion without altering the base model.
  \item We demonstrate that operating in a \emph{compact 128-D GFV space} enables efficient and effective RL training, with per-category specialization yielding better fidelity than a joint approach.
  \item We empirically validate our method on ShapeNetCore-2048 (5 categories), showing consistent improvements in Chamfer Distance (CD-L2) and F-Score@1\%, confirming RL-AD-Net’s effectiveness as a lightweight and general post-processing module.
\end{itemize}

\section{Related Work}

\subsection{Point Cloud Completion}
The task of point cloud completion aims to recover the full three-dimensional geometry of an object from partial or occluded observations. Early approaches were based on volumetric convolutional networks operating on voxel grids~\cite{dai2017shape, wang2021vepcn}. While effective for coarse shape recovery, these methods were constrained by the cubic growth in memory and computation with increasing resolution. Later, methods such as PCN~\cite{yuan2018pcn}, built upon PointNet-style encoders, directly processed unordered point sets and improved both scalability and reconstruction quality. More recently, transformer-based architectures such as AdaPoinTr~\cite{yu2022adapointr} have leveraged global attention mechanisms to capture long-range dependencies, leading to more coherent and detailed completions. In parallel, denoising-based approaches~\cite{lyu2021softpoolnet, zhou2022seedformer, zhang2023pointdiffusion} have emerged as another state-of-the-art direction, formulating completion as a progressive denoising process from noisy or partial point distributions. Despite this progress, even strong baselines often generate outputs with local geometric artifacts or missing regions, motivating the need for refinement strategies that can improve completion quality without retraining the base model.

\subsection{Latent Space Shape Representations}
Learning compact latent representations of three-dimensional shapes provides an efficient means for reconstruction, interpolation, and manipulation. Point cloud autoencoders (AEs)~\cite{achlioptas2018learning} and their variants map point sets to global feature vectors (GFVs) that capture semantic and structural properties of objects. These latent codes can be either category-specific or shared across multiple classes. Latent space refinement has been successfully applied in other domains such as image generation~\cite{wu2020stylespace} and mesh editing~\cite{gao2022clipmesh}, but its use in point cloud completion remains relatively unexplored. Operating in latent space offers a lightweight, model-agnostic interface that can serve as an effective target for refinement modules such as reinforcement learning agents, while keeping the base generator unchanged.

\subsection{Reinforcement Learning for Three-Dimensional Geometry Processing}
Reinforcement learning has been explored in several three-dimensional vision tasks including active object scanning~\cite{mousavian2016deep}, robotic grasping~\cite{qi2016pointnetgrasp}, and assembly planning~\cite{xu2022rlassembly}. For point cloud completion, RL-GAN-Net~\cite{sarmad2019rl} introduced an agent to guide the generation process of a GAN-based model. However, this required retraining both the generator and the discriminator, making the approach computationally demanding. More recently, Point-Patch RL~\cite{gyenes2024pointpatchrl} studied policy learning for completing local shape patches, but remained tightly coupled to a specific completion architecture. In contrast, our approach introduces a post-hoc latent refinement stage, where an RL agent operates exclusively in the latent space of a pre-trained autoencoder. This design allows seamless integration with different completion backbones without any modification to their training pipelines.

\subsection{Category-Specific Completion Strategies}
A number of studies have shown that models trained in a category-specific manner can outperform category-agnostic ones in classification~\cite{qi2017pointnetplusplus}, segmentation~\cite{wang2019dynamic}, and reconstruction~\cite{liu2020categoryspecific}. This improvement is attributed to the fact that each object category follows distinct geometric priors. For example, airplanes typically have symmetric elongated wings, while lamps often consist of thin vertical structures. In our work, we observed empirically that joint refinement across multiple categories degraded performance, likely due to conflicting geometric patterns in the shared latent space. These findings motivated the adoption of a category-wise reinforcement learning strategy, where dedicated agents are trained per class. This design choice is also consistent with many real-world scenarios where only a subset of categories is relevant for downstream applications, such as robotic bin-picking or autonomous driving perception.

\section{Proposed Framework}

Our proposed framework, \textbf{RL-AD-Net}, is a category-specific latent-space refinement pipeline that improves the output of any pre-trained point cloud completion network without modifying its weights. The method is structured in four main stages: (1) baseline completion using any point cloud completion network, (2) latent encoding using a category-specific autoencoder trained on Complete shapes, (3) reinforcement learning-based refinement of latent codes, and (4) geometry-aware selective refinement for deployment scenarios.

\begin{figure*}[!htb]
\centering
\includegraphics[width=0.95\textwidth]{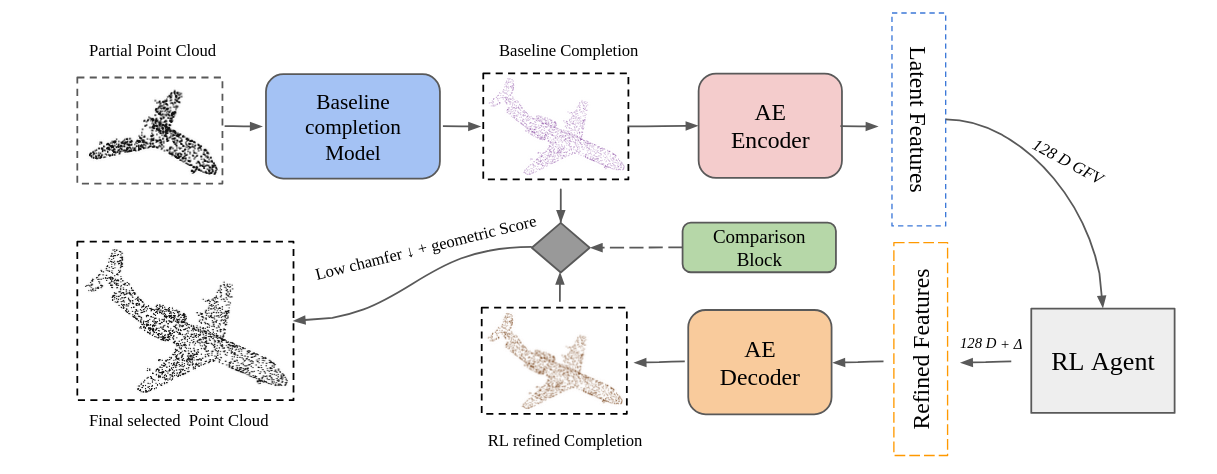}
\caption{\textbf{Overview of RL-AD-Net.}  
A partial point cloud is first completed by a pretrained baseline backbone.
The completion is encoded into a 128D Global Feature Vector (GFV) via the autoencoder.
A TD3 RL agent predicts an adjustment vector to refine the GFV.
The refined GFV is decoded back into a point cloud using the AE’s decoder.
A reward signal, based on Chamfer Distance improvement and action magnitude penalty, guides policy learning.Finally, a comparison block evaluates both the baseline and refined completions using Chamfer Distance and geometric consistency scores, and selects the better output as the final prediction.}
\label{fig:framework}
\end{figure*}

Figure~\ref{fig:framework} provides an overview.

\subsection{Baseline Completion with Any Point Cloud Network}

We adopt a model-agnostic approach where any pre-trained point cloud completion network can serve as the baseline generator. In our experiments, we primarily use AdaPoinTr~\cite{yu2022adapointr} due to its strong global-context modeling ability enabled by transformer-based attention, but our framework is equally applicable to other completion methods such as PCN~\cite{yuan2018pcn}, TopNet~\cite{tchapmi2019topnet}, or RGB2PC~\cite{lee2025rgb2point}. Given a partial point cloud $P_{\text{in}} \in \mathbb{R}^{N \times 3}$, the baseline network predicts a dense completion $P_{\text{base}} \in \mathbb{R}^{M \times 3}$.

\subsection{Complete-Shape Autoencoder Training}

To provide a compact latent space for refinement, we train a category-specific autoencoder (AE) solely on complete shapes from ShapeNetCore-2048. The AE consists of a PointNet-style encoder $E_\theta$ and a decoder $D_\phi$. Given a complete shape $P$, the encoder produces a global feature vector (GFV) $z \in \mathbb{R}^{128}$, which is decoded back into a reconstruction $\hat{P}$. Training minimizes the Chamfer Distance (CD) between $\hat{P}$ and $P$.

Unlike conventional completion models that are trained on partial-to-complete mappings, the AE is trained exclusively on complete shapes to learn a clean and unbiased geometric manifold. This ensures that the latent space represents the intrinsic structure of each category rather than artifacts induced by partial inputs or baseline completions . During refinement, the RL agent adjusts the latent codes of baseline completions obtained from this pretrained AE so that they better align with this well-formed manifold. This design allows the refinement process to exploit category-specific geometric priors, correcting local inaccuracies while maintaining global coherence.

\textbf{PointNet-Style Architecture Selection.} Our choice of PointNet-based architecture for the autoencoder is driven by three considerations: (i) lightweight design ensures minimal computational overhead, enabling efficient training and fast RL policy evaluation; (ii) the 128-dimensional latent representation balances capturing category-specific geometric information with a tractable action space for RL convergence; and (iii) architectural simplicity ensures broad compatibility with diverse completion models.
We train the AE for 400 epochs using the Adam optimizer with learning rate $0.001$, momentum $0.9$, and $\beta = 0.999$. The batch size is 24, and a multi-step scheduler reduces the learning rate at epochs $\{60, 120, 180, 400\}$ with decay $\gamma = 0.5$. Both encoder and decoder are trained end-to-end with Chamfer Distance loss. After training, the decoder $D_\phi$ is frozen and reused for decoding refined latent vectors.

\subsection{GFV Generation from Baseline Completions}

Once the AE is trained, we generate GFVs for RL training by passing baseline completions through the AE encoder. Specifically, given $P_{\text{base}}$ from any completion network, the encoder produces $z = E_\theta(P_{\text{base}})$. The resulting GFVs for the state space for RL. These are stored offline for each category, creating a dataset of state ground truth pairs used to train our category-specific RL agent.

\subsection{Reinforcement Learning for Latent Refinement}

For each category $c$, we train a separate RL agent $\pi^{(c)}_\psi$ to refine latent vectors. The refinement is formulated as a continuous control task:

\textbf{State.} The input state is the GFV $z \in \mathbb{R}^{128}$ obtained from the AE encoder.

\textbf{Action.} The action is a refinement vector $\Delta z \in \mathbb{R}^{128}$ predicted by the RL policy. The refined code is:
\begin{equation}
    z' = z + \alpha \Delta z,
\end{equation}
where $\alpha$ is a refinement scaling factor.

\textbf{Reward.} The reward encourages improvements in Chamfer Distance (CD) between the reconstructed completion and the ground truth:
\begin{equation}
    R = CD(P_{\text{base}}, P_{\text{gt}}) - CD(P_{\text{ref}}, P_{\text{gt}}),
\end{equation}
where $P_{\text{ref}} = D_\phi(z')$. Positive rewards are given when refinement reduces reconstruction error.

\textbf{Policy Optimization.} We adopt Twin Delayed Deep Deterministic Policy Gradient (TD3)~\cite{fujimoto2018td3}, which is well-suited for high-dimensional continuous control. The actor learns to propose refinement vectors, while twin critics stabilize value estimation. Training proceeds over stored GFVs with mini-batches of size 64, target smoothing noise, and delayed policy updates to mitigate overestimation bias.

\subsection{Geometry-Aware Selective Refinement for Deployment}

At inference, ground truth is typically unavailable, making it necessary to assess the quality of completions without direct supervision. To this end, we employ \textbf{PointNN}~\cite{zhang2023parameter}, a non-parametric network that evaluates geometric consistency of point clouds. Given a baseline completion $P_{\text{base}}$ and an RL-refined output $P_{\text{ref}}$, PointNN assigns a quality score $q \in [0,1]$ to each. The final output is selected as:
\[
P_{\text{out}} = 
\begin{cases} 
P_{\text{ref}}, & q_{\text{ref}} > q_{\text{base}}, \\ 
P_{\text{base}}, & \text{otherwise.}
\end{cases}
\]

This ensures that refinement never degrades performance, since the system always chooses the geometrically superior completion. In evaluation scenarios where ground truth $P_{\text{gt}}$ is available, we further combine the PointNN score with Chamfer Distance (CD), selecting the completion that achieves both lower CD and higher geometric consistency. This dual criterion provides stronger validation while maintaining robustness in deployment without ground truth.

\section{Experiments}

We evaluate the proposed \textbf{RL-AD-Net} framework on the ShapeNetCore-2048 dataset under a category-wise setting. The experiments are designed to quantify the improvement in reconstruction quality brought by category-specific RL refinement over a strong transformer-based completion baseline.For Baseline model training step we used \textit{AdaPoinTr}~\cite{yu2022adapointr} model for our experiment and it is trained for $300$ epochs over Chamfer Distance loss.

\subsection{Dataset and Preprocessing}
For generation of partial inputs we used two types of occlusion strategies:

\begin{itemize}
    \item \textbf{Spherical region-based cropping:} Given an input point cloud $P \in \mathbb{R}^{2048 \times 3}$, a random unit vector $c \in \mathbb{R}^{1 \times 3}$ is sampled and used as a reference direction. Points are ranked by their Euclidean distance to $c$, and the closest $k$ points are removed to achieve fixed occlusion levels (e.g., $25\%$ or $50\%$). This strategy mimics view-dependent occlusions commonly used in transformer-based methods such as AdaPoinTr~\cite{yu2022adapointr}.  

    \item \textbf{Seed-point proximity removal:} A random seed point is selected from the point cloud, and squared distances of all other points to this seed are computed. Points within the smallest $d\%$ radius are removed, where $d$ is the deletion ratio (set to $40\%$ in our experiments). This results in partial clouds with approximately $1229$ points, effectively simulating local occlusions centered around arbitrary object regions.  
\end{itemize}

We conduct experiments on five representative categories from ShapeNet (\emph{Airplane, Chair, Lamp, Table, Car}), training separate autoencoders (AEs) and RL agents for each category.

\subsection{Evaluation Metrics}
Reconstruction quality is assessed using two widely adopted metrics:  
\begin{itemize}
    \item \textbf{Chamfer Distance (CD-L2):} the average squared nearest-neighbor distance between prediction and ground truth point sets (lower is better).  
    \item \textbf{F-Score@1\%:} the harmonic mean of precision and recall under a tolerance of $1\%$ of the object’s bounding box diagonal (higher is better).  
\end{itemize}

\subsection{Implementation Details}
The autoencoder uses a PointNet-style encoder that maps each input to a $128$-dimensional global feature vector (GFV), and a decoder that reconstructs $2048$ points from this latent code. The RL agent operates in this 128-dimensional latent space, outputting an additive refinement vector.  

Each AE is trained for $400$ epochs per category with Adam optimizer (learning rate $1\mathrm{e}{-4}$, batch size $32$) using Chamfer Distance loss.The RL agent is optimized using TD3~\cite{fujimoto2018td3} for \textbf{100,000 training iterations}, with a replay buffer size of $10^5$, policy delay $2$, target smoothing coefficient $\tau = 0.005$, and Gaussian exploration noise $\sigma = 0.1$.

\section{Results}

\subsection{Spherical Region-Based Cropping (25\% and 50\%)}
We first evaluate RL-AD-Net under the spherical region-based cropping experiment. Results are reported separately for Chamfer Distance and F-Score.  

The quantitative results demonstrate consistent improvements across all categories under both 25\% and 50\% cropping scenarios. For Chamfer Distance (Table~\ref{tab:cd_results}), our method achieves notable reductions, with the most significant improvement observed in the \emph{Airplane} category where CD-L2 decreases from 1.183 to 1.159 under 25\% cropping. Similarly, F-Score@1\% results (Table~\ref{tab:fscore_results}) show improvements across categories, with \emph{Airplane} again showing substantial gains from 0.329 to 0.420 under 25\% cropping conditions. These improvements persist even under the more challenging 50\% crop, where RL-AD-Net achieves better reconstruction fidelity compared to AdaPoinTr.  

\begin{table}[h]
\centering
\caption{Chamfer Distance (CD-L2) on ShapeNetCore-2048 under spherical region-based 25\% and 50\% cropping (lower is better).}
\label{tab:cd_results}
\begin{tabular}{lccccc}
\hline
 & Airplane & Chair & Lamp & Table & Car \\
\hline
25\% AdaPoinTr & 1.183  & 1.472 & 1.296 & 1.392 & 1.582 \\
25\% RL-AD-Net  & \textbf{1.159}  & \textbf{1.470} & \textbf{1.272} & \textbf{1.386} & \textbf{1.575} \\
50\% AdaPoinTr & 1.321  & 1.663 & 1.588 & 1.500 & 1.662 \\
50\% RL-AD-Net  & \textbf{1.235}  & \textbf{1.655} & \textbf{1.582} & \textbf{1.470} & \textbf{1.648} \\
\hline
\end{tabular}
\end{table}

\begin{table}[h]
\centering
\caption{F-Score@1\% on ShapeNetCore-2048 under spherical region-based 25\% and 50\% cropping (higher is better).}
\label{tab:fscore_results}
\begin{tabular}{lccccc}
\hline
 & Airplane & Chair & Lamp & Table & Car \\
\hline
25\% AdaPoinTr & 0.329 & 0.130 & 0.272 & 0.147 & 0.124 \\
25\% RL-AD-Net  & \textbf{0.420} & \textbf{0.131} & \textbf{0.278} & \textbf{0.150} & \textbf{0.131} \\
50\% AdaPoinTr & 0.236 & 0.072 & 0.170 & 0.118 & 0.071 \\
50\% RL-AD-Net  & \textbf{0.348} & \textbf{0.078} & \textbf{0.172} & \textbf{0.122} & \textbf{0.087} \\
\hline
\end{tabular}
\end{table}

\begin{figure*}[!htb]
  \centering
  \includegraphics[width=0.95\textwidth]{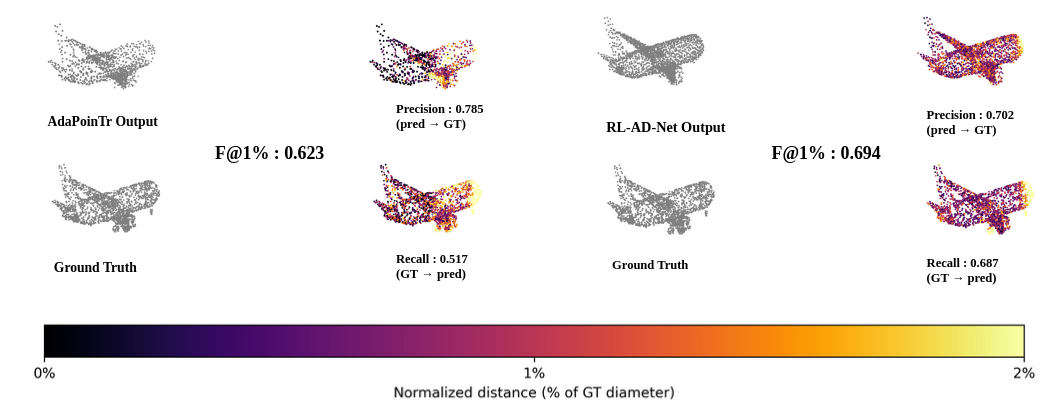}
  \caption{Comparison of AdaPoinTr and RL-AD-Net outputs on \emph{Airplane} (50\% crop). Left: AdaPoinTr baseline; Right: RL-AD-Net refinement. RL-AD-Net improves recall while preserving precision, leading to a higher F-score@1\%.}
  \label{fig:qual_airplane}
\end{figure*}

\begin{figure*}[!htb]
\centering
\includegraphics[width=0.95\textwidth]{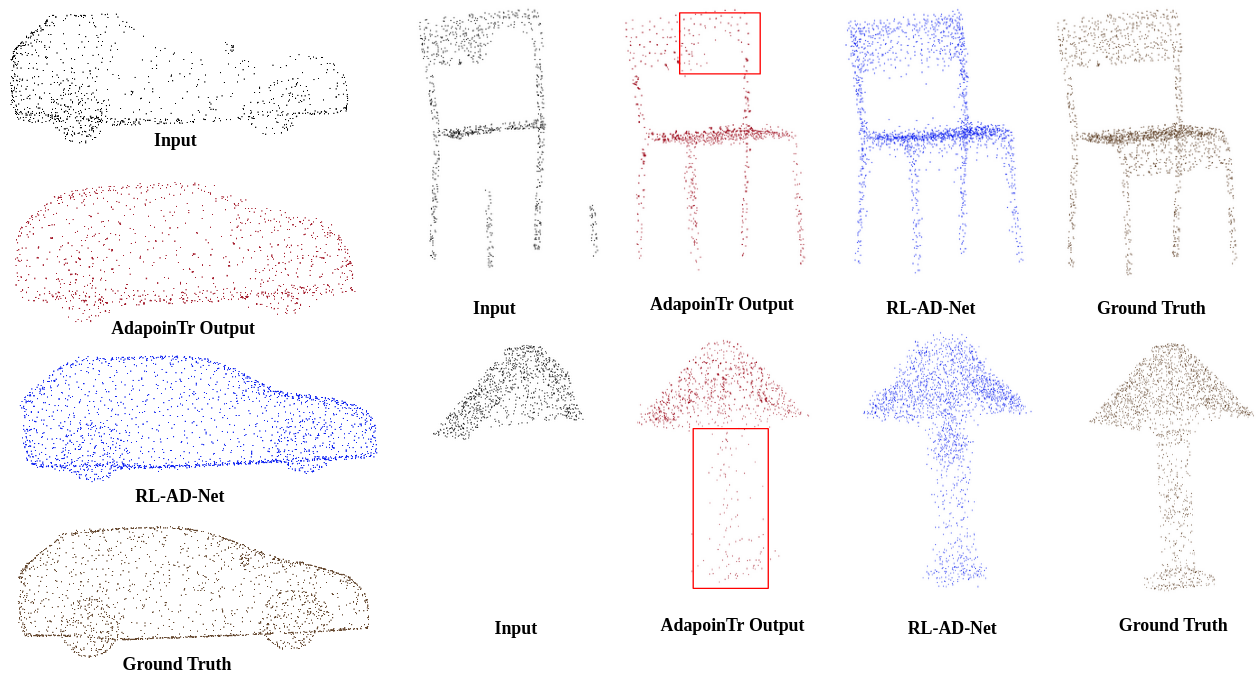}
\caption{Qualitative visualizations under \textbf{50\% spherical cropping}. Categories: \emph{Car}, \emph{Chair}, and \emph{Lamp}. RL-AD-Net reduces structural gaps and recovers missing regions shown in red squares while preserving fine details.}
\label{fig:qualitative_results2}
\end{figure*}

The visual comparisons further confirm these improvements. Figure~\ref{fig:qual_airplane} presents a representative case study for the \emph{Airplane} category under 50\% cropping. In this instance, RL-AD-Net improves recall from 0.517 to 0.687 while maintaining comparable precision, leading to a higher overall F-score@1\%. These values are shown to illustrate the effect on a single example, whereas the quantitative tables report category-level averages. Figures~\ref{fig:qualitative_results1} and~\ref{fig:qualitative_results2} provide additional qualitative comparisons across categories. Under 25\% cropping (Figure~\ref{fig:qualitative_results1}), RL-AD-Net yields more complete reconstructions for \emph{Car}, \emph{Airplane}, \emph{Table}, and \emph{Lamp}. Under 50\% cropping (Figure~\ref{fig:qualitative_results2}), it reduces structural gaps and recovers missing regions in \emph{Car}, \emph{Chair}, and \emph{Lamp}. These examples visually illustrate the improvements, complementing the quantitative results in the tables.

\begin{figure*}[!htb]
\centering
\includegraphics[width=0.95\textwidth]{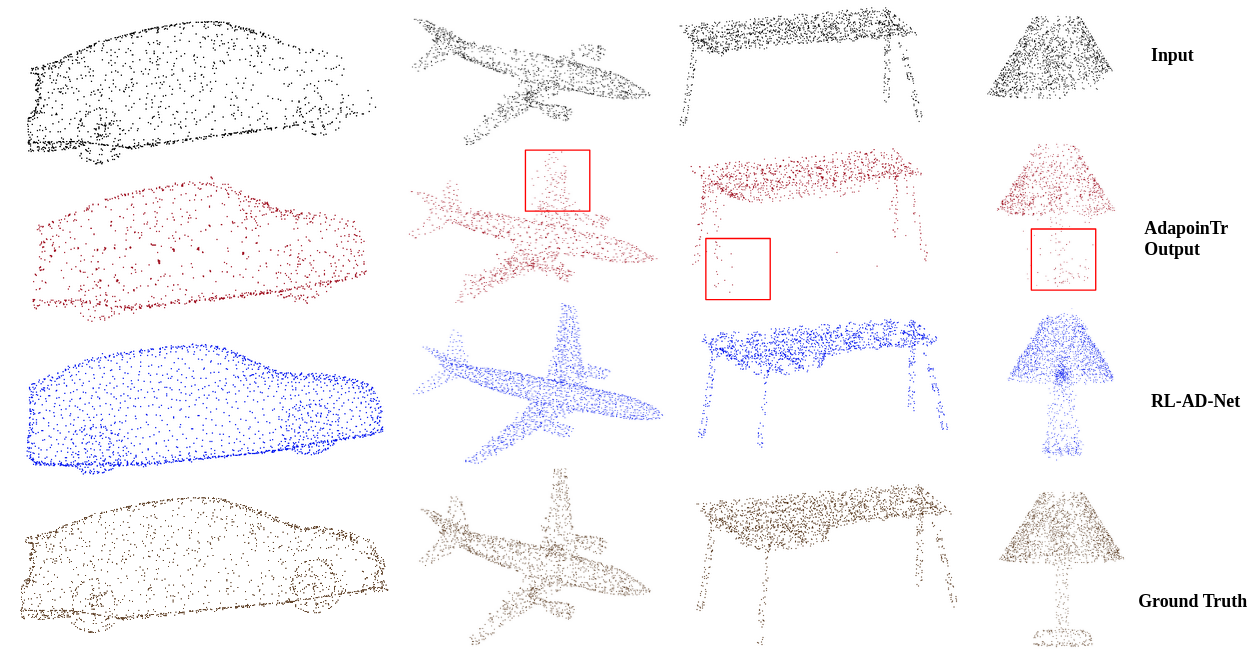}
\caption{Qualitative visualizations under \textbf{25\% spherical cropping}. Categories: \emph{Car}, \emph{Airplane}, \emph{Table}, and \emph{Lamp}. RL-AD-Net yields more complete and geometrically consistent reconstructions than AdaPoinTr as marked with red squares.}
\label{fig:qualitative_results1}
\end{figure*}

\textbf{Runtime and Latent-Space Analysis}
Table~\ref{tab:policy_complexity} reports the size and inference cost of the TD3 policy used in RL-AD-Net. The actor has only 0.210M parameters and 0.210M multiply–accumulate operations (MACs) with a mean latency of 0.1036ms, while the critic has 0.447M parameters and 0.445M MACs at 0.1801ms. Overall, the learned RL head remains sub-millisecond and adds negligible overhead relative to completion and metric computation.

\begin{table}[t]
  \centering
  \caption{Policy network complexity and latency (batch = 1). 
  \textbf{Params (M)}: number of trainable parameters in millions. 
  \textbf{MACs (M)}: multiply accumulate operations in millions, estimated via \texttt{thop}. 
  \textbf{Mean / p50 / p90 / p95 (ms)}: average, median, 90th percentile, and 95th percentile inference latency (milliseconds) measured over 400 runs with 100 warmup iterations. 
  \textbf{Ckpt (MB)}: checkpoint size in megabytes. 
  Hardware: NVIDIA GeForce RTX 4060 Ti, FP32.}
  \label{tab:policy_complexity}
  \begin{tabular}{lcc}
    \toprule
    \textbf{Metric} & \textbf{Actor} & \textbf{Critic} \\
    \midrule
    Params (M)   & 0.210  & 0.447  \\
    MACs (M)     & 0.210  & 0.445  \\
    Mean (ms)    & 0.1036 & 0.1801 \\
    p50 (ms)     & 0.1048 & 0.1464 \\
    p90 (ms)     & 0.1088 & 0.1533 \\
    p95 (ms)     & 0.1109 & 0.1556 \\
    Ckpt (MB)    & 0.84   & 1.79   \\
    \bottomrule
  \end{tabular}
\end{table}

\subsection{Seed-point proximity removal Cropping}
To further evaluate robustness, we consider a more challenging 40\% random cropping setup. Unlike spherical crops, random masking can disrupt object symmetries and remove structurally critical regions, making completion more difficult. As before, we compare the baseline AdaPoinTr completions with the final outputs after selective RL refinement, where the system chooses per sample between the original and refined predictions.  

The quantitative results in Table~\ref{tab:cd40} show consistent Chamfer Distance reductions across categories. The largest gains are observed in structurally diverse categories such as \emph{Chair}, \emph{Lamp}, and \emph{Table}, with CD-L2 decreasing by nearly one full unit on average (e.g., Chair from 5.574 to 4.658). Even in the more compact \emph{Car} category, RL-AD-Net achieves a noticeable improvement from 5.340 to 4.518.  

\begin{table}[h]
\centering
\caption{Chamfer Distance (CD-L2) under 40\% random cropping on ShapeNetCore-2048 (lower is better).}
\label{tab:cd40}
\begin{tabular}{lcc}
\hline
Category & AdaPoinTr & RL-AD-Net \\
\hline
Airplane & 4.283 & \textbf{3.159} \\
Chair    & 5.574 & \textbf{4.658} \\
Lamp     & 6.553 & \textbf{5.703} \\
Table    & 6.097 & \textbf{5.325} \\
Car      & 5.340 & \textbf{4.518} \\
\hline
\end{tabular}
\end{table}

The F-Score results in Table~\ref{tab:fscore40} further highlight the advantage of selective RL refinement. \emph{Airplane} benefits substantially, improving from 0.221 to 0.279, while categories with irregular structures such as \emph{Chair} and \emph{Car} nearly double their F-score@1\% scores. Interestingly, \emph{Table} shows stable performance (0.090 to 0.117), suggesting that refinement is conservative when structural confidence is already high.  

\begin{table}[h]
\centering
\caption{F-Score@1\% under 40\% random cropping on ShapeNetCore-2048 (higher is better).}
\label{tab:fscore40}
\begin{tabular}{lcc}
\hline
Category & AdaPoinTr  & RL-AD-Net  \\
\hline
Airplane & 0.221 & \textbf{0.279} \\
Chair    & 0.058 & \textbf{0.096} \\
Lamp     & 0.127 & \textbf{0.154} \\
Table    & 0.090 & \textbf{0.117} \\
Car      & 0.052 & \textbf{0.076} \\
\hline
\end{tabular}
\end{table}

Overall, the random cropping results confirm that RL-AD-Net not only improves completion accuracy in regular categories like \emph{Airplane}, but also adapts effectively to irregular and structurally diverse objects. The per-sample selection strategy ensures that improvements are achieved without destabilizing reliable baseline cases, highlighting the generality of the refinement process across different masking patterns.

\section{Conclusion}

We presented \textbf{RL-AD-Net}, a reinforcement learning-based post-processing framework for category-specific point cloud completion refinement.  
The method operates directly in the latent space of a compact PointAE autoencoder, enabling efficient geometry-aware adjustments to Baseline completions without retraining the base model.  

Our experiments on ShapeNetCore-2048 reveal that our Baseline performs reasonably well under its own training-style cropping, but struggles in random cropping settings. In contrast, our \textbf{RL-AD-Net} consistently outperforms Baseline across both spherical and random cropping protocols. By leveraging an ensemble-style refinement strategy selecting per instance between baseline and RL-refined outputs the method achieves systematic improvements in Chamfer Distance and F-Score across diverse conditions.  

The robustness of this selective refinement highlights reinforcement learning as a powerful ensemble mechanism for point cloud completion. RL-AD-Net is broadly applicable to any latent-encoding completion network, with category-specific specialization reflecting realistic deployment needs. As future work, we aim to extend the approach toward cross-category refinement, more expressive encoders, and dynamic point cloud streams.



\section*{Supplementary Material}
\addcontentsline{toc}{section}{Supplementary Material}

\subsection*{Contents}
\begin{enumerate}[leftmargin=1.2em,itemsep=0.2em]
  \item Random 40\% partial input results (visual \& quantitative)
  \item Magnified views of t\textsc{-}SNE latent trajectories
  \item Single-image RGB$\rightarrow$Point Cloud completion refined by RL-AD-Net
  \item Policy training curves: PPO vs.\ TD3 vs.\ DDPG
  \item Multi-category aggregate metrics (5 categories)
  \item Miscellaneous: Point-NN for Geometry-Aware Selection
\end{enumerate}

\section{Random 40\% Partial Input Results}
\label{sec:supp_random40}

We remove 40\% of points using seed-point proximity masking (see main paper for details). For each sample, the final prediction is chosen via our geometry-aware selector combining Chamfer distance and geometric consistency. Figure~\ref{fig:qualitative_results1} shows representative qualitative results, while Table~\ref{tab:cd40} reports per-category Chamfer distances.

Although RL-AD-Net refines the AdaPoinTr completions and generally improves Chamfer distance metrics (Table~\ref{tab:cd40}), its performance remains bounded by the quality of the AdaPoinTr outputs. Since the refinement operates in the latent space of the AdaPoinTr completions, any structural errors or hallucinations introduced during the initial completion are inevitably propagated into the RL stage. This effect is particularly visible in the 40\%  random cropping setup, where AdaPoinTr already suffers higher Chamfer distance compared to the spherical cropping case. As a result, RL-AD-Net refinements inherit these degradations, even though they produce denser and smoother completions.  

Figure~\ref{fig:qualitative_results1} highlights these challenges. The black squares mark regions where AdaPoinTr introduces spurious structures inconsistent with the ground truth, effectively hallucinating geometry. The red circles indicate areas where RL-AD-Net, constrained by these flawed initial completions, fails to fully recover structural fidelity and in some cases even amplifies the artifacts. This illustrates the dependency of our refinement strategy on the upstream backbone while RL-AD-Net consistently reduces Chamfer distance relative to AdaPoinTr, its ultimate accuracy is bounded by the geometric plausibility of the baseline completions.

\begin{figure*}[!htb]
\centering
\includegraphics[width=0.95\textwidth]{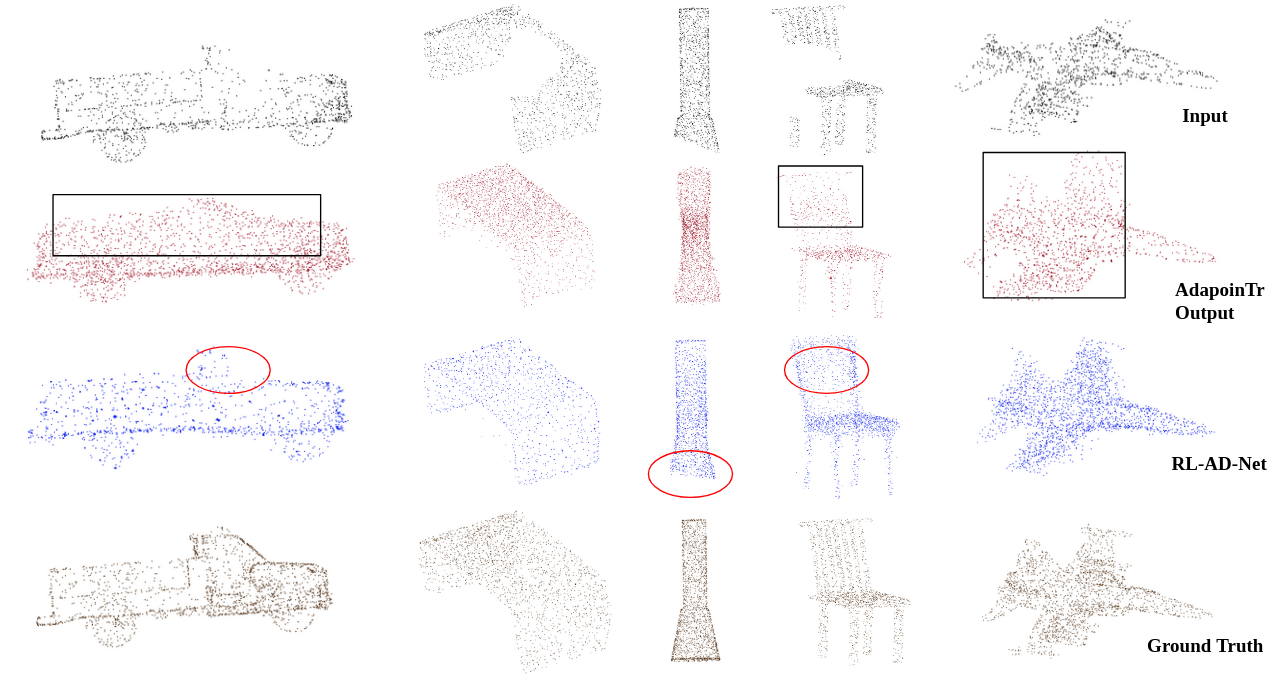}
\caption{Qualitative visualizations under \textbf{40\% random masking}. Categories: \emph{Car}, \emph{Table}, \emph{Lamp} ,\emph{chair} \ and \emph{Airplane}. Black boxes mark hallucinated or distorted regions in AdaPoinTr outputs relative to the ground truth. Red circles highlight refinement failures in RL-AD-Net caused by propagation of these errors.}
\label{fig:qualitative_results1}
\end{figure*}

\begin{table}[h]
\centering
\caption{Chamfer Distance (CD-L2) under 40\% random cropping on ShapeNetCore-2048 (lower is better).}
\label{tab:cd40}
\begin{tabular}{lcc}
\hline
Category & AdaPoinTr & RL-AD-Net \\
\hline
Airplane & 4.283 & \textbf{3.159} \\
Chair    & 5.574 & \textbf{4.658} \\
Lamp     & 6.553 & \textbf{5.703} \\
Table    & 6.097 & \textbf{5.325} \\
Car      & 5.340 & \textbf{4.518} \\
\hline
\end{tabular}
\end{table}

\section{Magnified t\textsc{-}SNE Latent Trajectories}
\label{sec:supp_tsne_zoom}



\begin{figure*}[t]
  \centering
  \begin{subfigure}{0.45\linewidth}
    \centering
    \includegraphics[width=\linewidth]{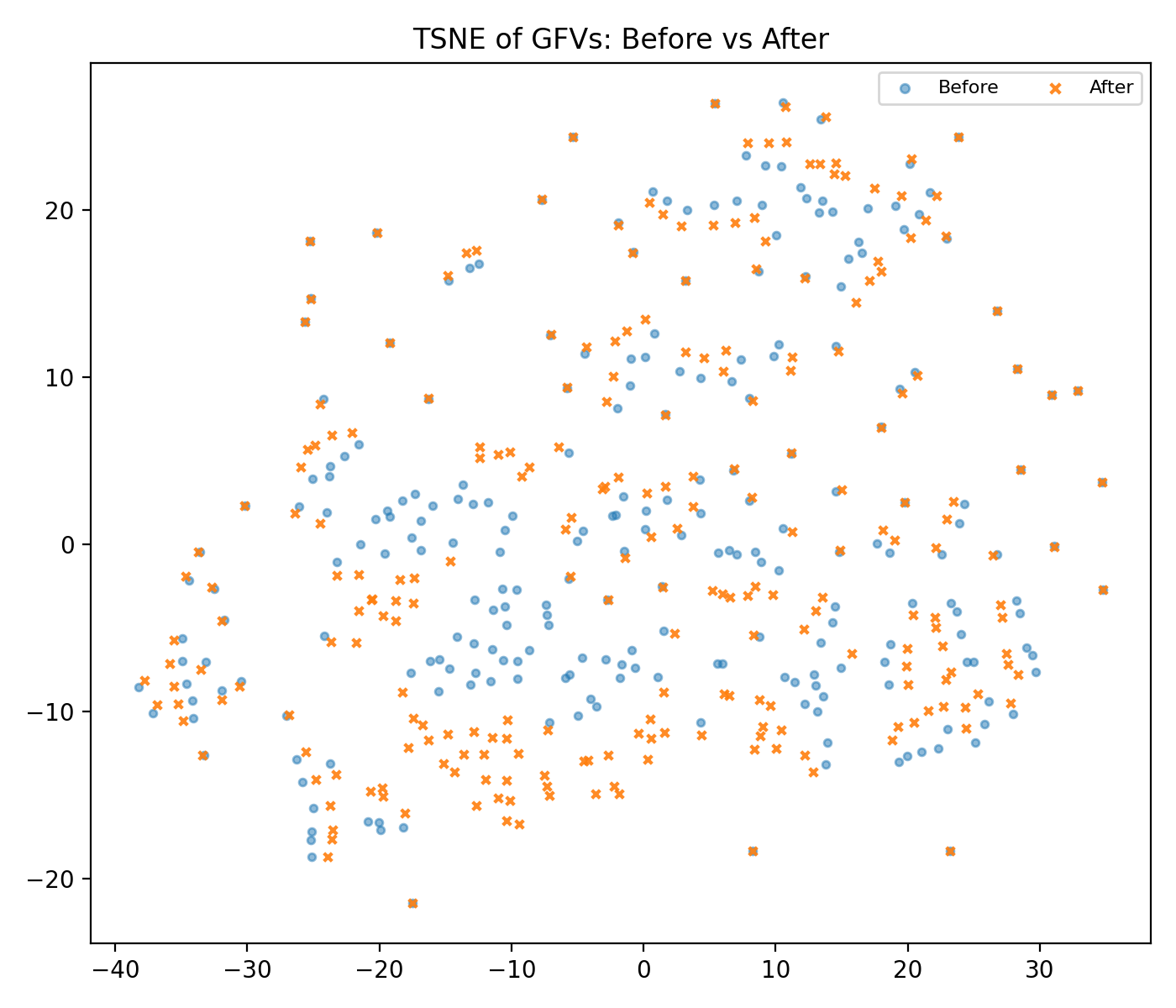}
    \caption{Airplane: before vs.\ after refinement.}
  \end{subfigure}\hfill
  \begin{subfigure}{0.45\linewidth}
    \centering
    \includegraphics[width=\linewidth]{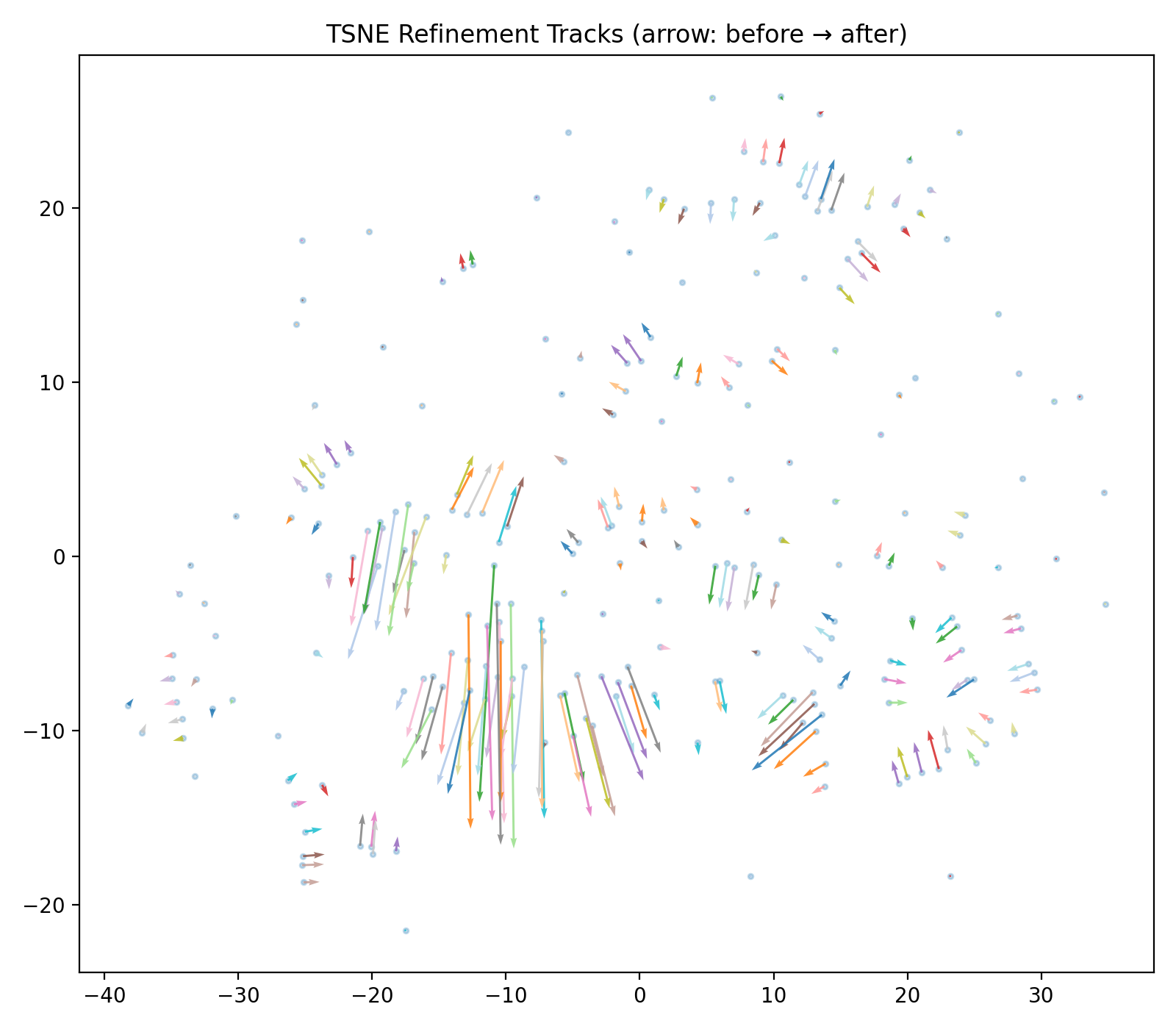}
    \caption{Airplane: latent trajectories with arrowed displacements.}
  \end{subfigure}
  \caption{\textbf{Magnified t\textsc{-}SNE views.}  
  Each arrow points from the baseline GFV to its RL-refined counterpart. The displacements are locally coherent, highlighting that RL-AD-Net performs geometry-aware corrections rather than random shifts.}
  \label{fig:supp_tsne_zoom}
\end{figure*}

We project AE latent codes (GFVs) to 2D with t\textsc{-}SNE and illustrate per-instance displacements from baseline to refined embeddings. While the main paper already presented these results, the visualizations appeared compressed due to space limitations. To aid interpretation, Figure~\ref{fig:supp_tsne_zoom} provides magnified views to make the latent trajectories clearer.  

The enlarged plots highlight how RL-AD-Net induces structured, locally coherent shifts in the latent space rather than random perturbations. Each arrow originates from the AdaPoinTr baseline embedding and terminates at the RL-refined counterpart, revealing smooth and geometry-aware corrections. These magnified t\textsc{-}SNE views confirm the policy’s ability to capture meaningful refinements in the latent manifold, consistent with the analysis in the main paper.

\section{RGB-to-Point Cloud Completion with RL Refinement}
\label{sec:supp_rgb2pc}

One of the strengths of RL-AD-Net lies in its plug-and-play design, allowing integration with different point cloud generation backbones.To demonstrate this, we replaced AdaPoinTr with a single-image to point-cloud generator RGB2Point model and ran it on the same ShapeNetCore rendered images on which it was trained to get the initial completions. Each completion is then encoded by our autoencoder (AE) to obtain a global feature vector (GFV), refined in latent space by the RL policy, and decoded back to a point cloud. Finally, a selection module compares the baseline and refined reconstructions using Chamfer Distance and a geometry-aware consistency score, and retains the superior result as the final prediction.

As shown in Figure~\ref{fig:qualitative_results1}, RL-AD-Net successfully improves the structural fidelity and completeness of RGB2Point outputs, validating its ability to operate as a refinement layer beyond AdaPoinTr. Importantly, we found that proper scale normalization and data preprocessing steps, consistent with those used in autoencoder (AE) training, are essential for such generative modules. Without normalization, the raw RGB2Point outputs may distort global feature vectors (GFVs), degrading both refinement quality and reconstruction stability. This underscores the necessity of standardized preprocessing when integrating diverse upstream modules. Overall, these results highlight RL-AD-Net’s generality that can refine completions from heterogeneous sources, provided consistent normalization is applied.

\begin{figure*}[!htb]
\centering
\includegraphics[width=0.95\textwidth]{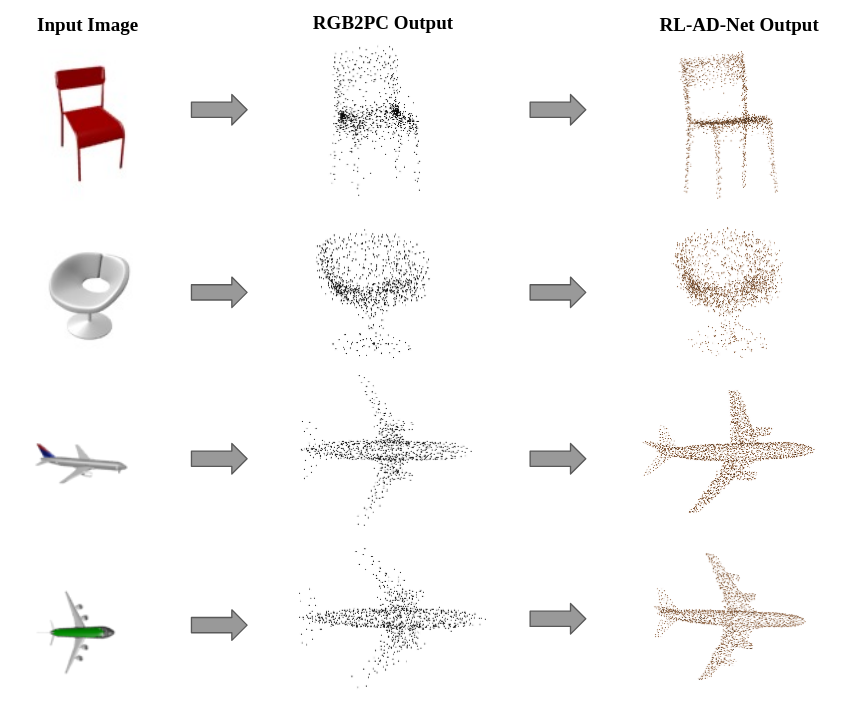}
\caption{\textbf{Plug-and-play refinement with RGB2Point.}  
Single-image point cloud completions from RGB2Point are passed through RL-AD-Net for latent-space refinement. The framework enhances geometric consistency and completion quality, provided that outputs are properly normalized before encoding. Red boxes mark regions where RL-AD-Net yields visibly improved reconstructions over the baseline RGB2Point outputs.}

\label{fig:qualitative_results1}
\end{figure*}

\section{Policy Training Curves: PPO, TD3, DDPG}
\label{sec:supp_policy_curves}

\begin{figure*}[t]
  \centering
  \includegraphics[width=0.9\linewidth]{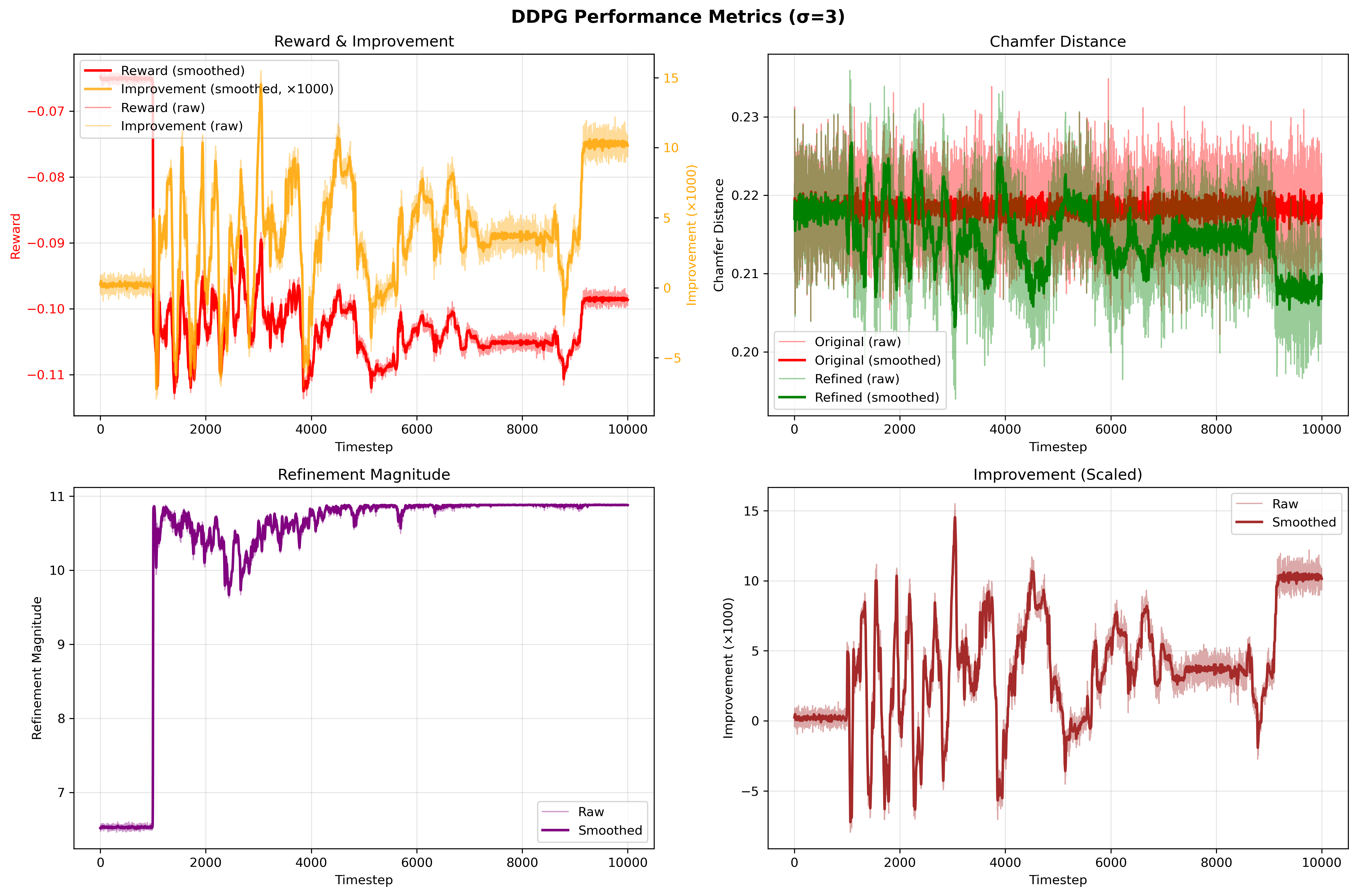}
  \caption{\textbf{DDPG performance metrics.} Reward progression, Chamfer distance reduction, refinement magnitude, and scaled improvement across 10k timesteps. DDPG achieves consistent Chamfer distance gains but shows moderate training variance.}
  \label{fig:ddpg_metrics}
\end{figure*}

\begin{figure*}[t]
  \centering
  \includegraphics[width=0.9\linewidth]{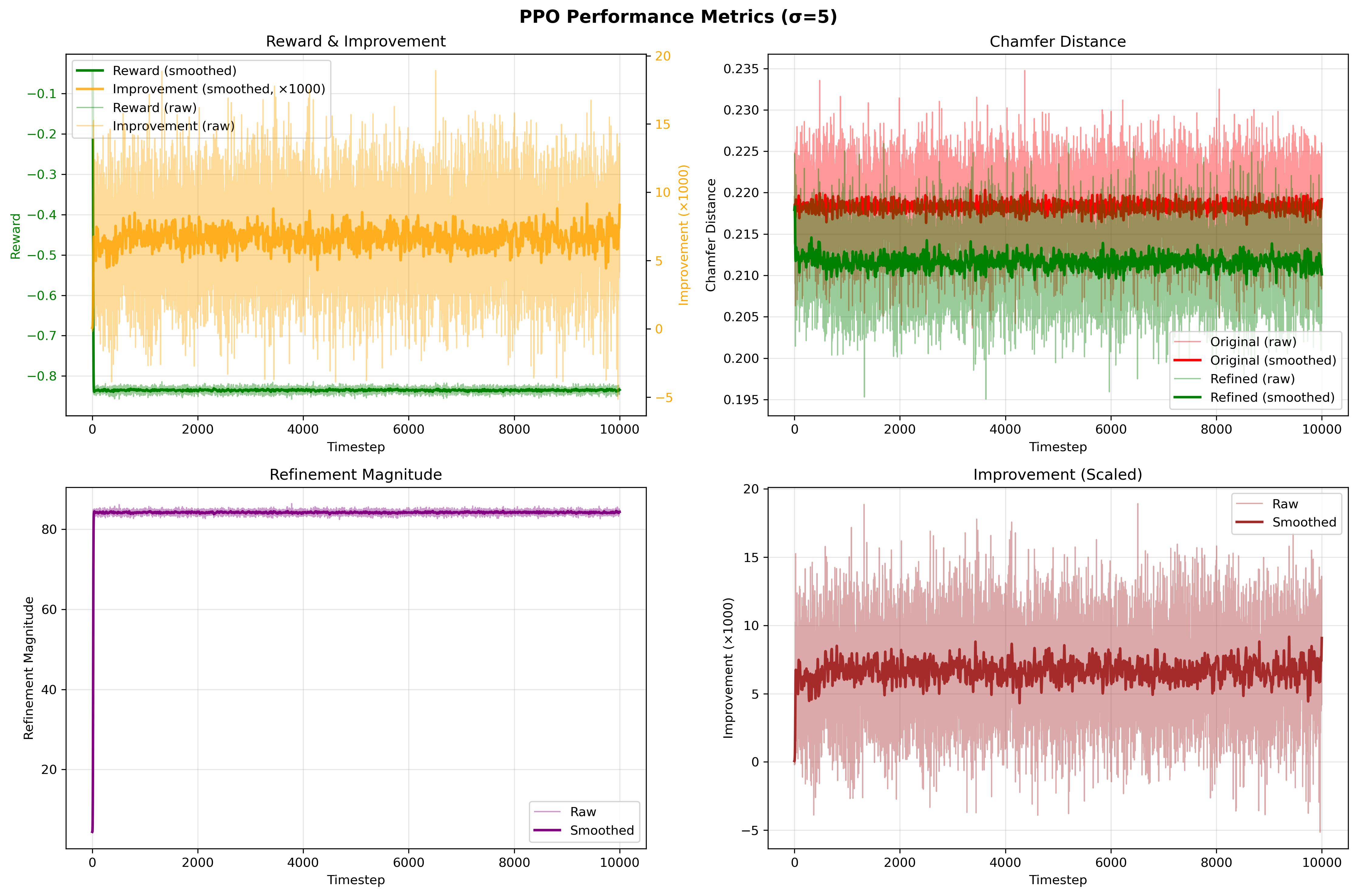}
  \caption{\textbf{PPO performance metrics.} Reward progression, Chamfer distance reduction, refinement magnitude, and scaled improvement across 10k timesteps. PPO exhibits poor reward learning, unstable refinement magnitudes, and minimal Chamfer distance improvement.}
  \label{fig:ppo_metrics}
\end{figure*}

\begin{figure*}[t]
  \centering
  \includegraphics[width=0.9\linewidth]{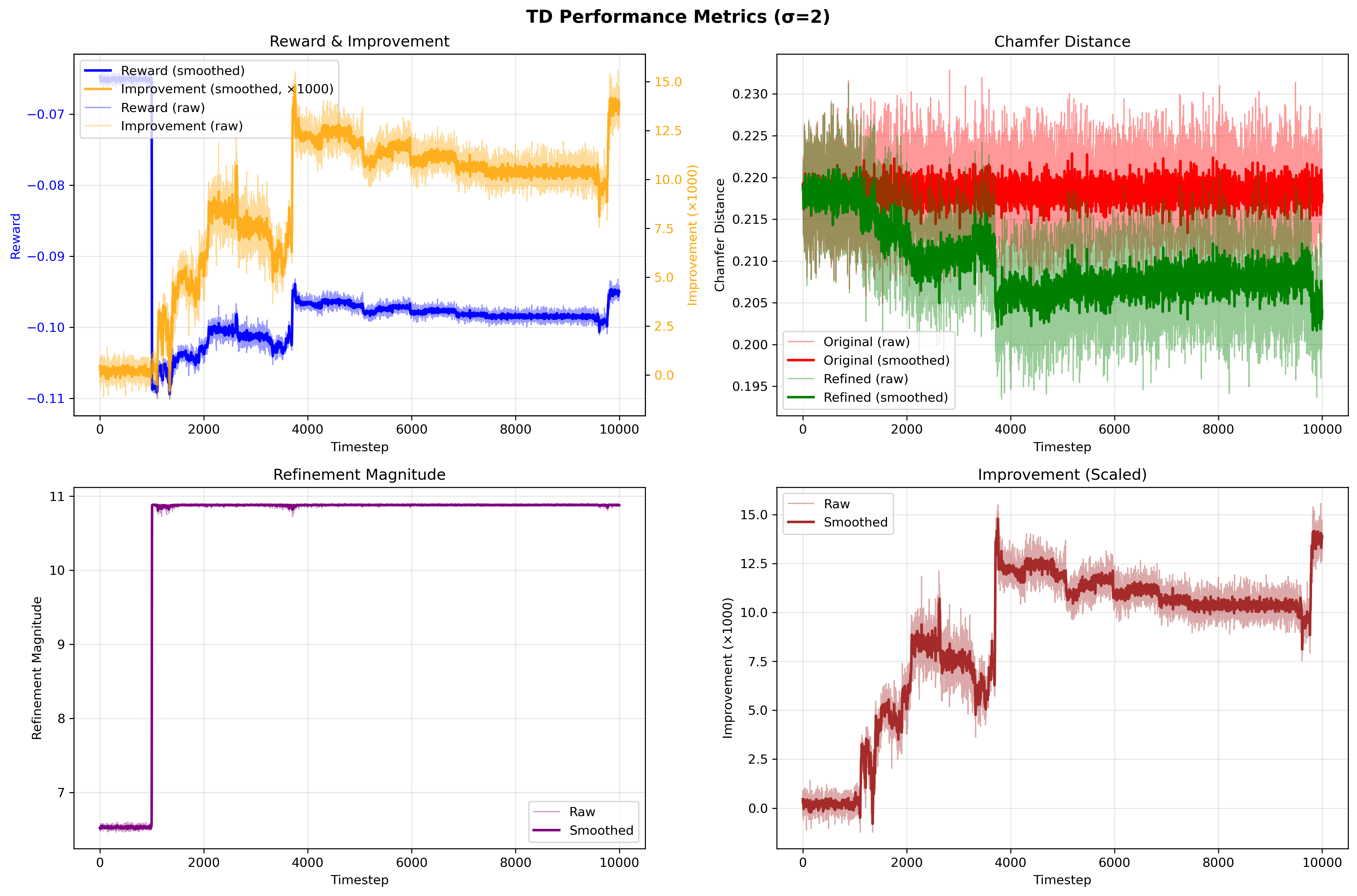}
  \caption{\textbf{TD performance metrics.} Reward progression, Chamfer distance reduction, refinement magnitude, and scaled improvement across 10k timesteps. TD demonstrates the best learning stability, strongest Chamfer distance reduction, and smooth convergence.}
  \label{fig:td_metrics}
\end{figure*}

The comparative evaluation of three reinforcement learning algorithms TD3, DDPG, and PPO for the refinement stage of RL-AD-Net is summarized in Figures~\ref{fig:ddpg_metrics}--\ref{fig:td_metrics}. TD3 (Figure~\ref{fig:td_metrics}) shows the most favorable performance, with rewards progressing steadily from approximately $-0.11$ to $-0.095$ and Chamfer distance reductions that consistently separate refined outputs from their baselines. Its refinement magnitude stabilizes around 11, indicating controlled policy updates, while the improvement curves demonstrate clear two-phase learning with strong convergence. By contrast, DDPG (Figure~\ref{fig:ddpg_metrics}) achieves moderate improvements with final rewards near $-0.10$ and visible Chamfer distance gains, but training variance remains higher and learning less smooth than TD3. Despite this volatility, DDPG still demonstrates effective refinement behavior, maintaining refinement magnitudes in a stable range and achieving steady gains in the improvement metric.

In contrast, PPO (Figure~\ref{fig:ppo_metrics}) fails to achieve meaningful progress under the same reward and environment setup. Rewards remain flat around $-0.82$, refined Chamfer distances show little separation from the baseline, and refinement magnitudes escalate to excessively high values ($\sim 85$), indicating unstable and inefficient updates. This divergence underscores the poor suitability of PPO for geometry-aware latent space refinement, where stability and fine-grained policy control are critical. Overall, the figure-based evidence confirms that TD3 is the most effective choice, combining fast convergence, low volatility, and strong geometric improvements, followed by DDPG with acceptable but less stable performance, while PPO lags far behind in both learning efficiency and refinement effectiveness.

\section{Multi-Category RL training Results}
\label{sec:supp_multi_category}

We also report the combined category trained RL and AE model performance with per-category sample counts. Metrics are computed using the per-sample selection between baseline and refined predictions.

\begin{table}[t]
  \centering
  \caption{Multi-category RL training: per-category test results on ShapeNetCore-2048.
  Baseline is the completion backbone alone; \emph{Final} applies per-sample selection between the backbone and RL-refined output.
  Under joint (multi-class) RL training, the number of selected refined samples is negligible (3--6 per category out of 800+), yielding only marginal differences in final metrics.}
  \label{tab:supp_multicat_results}
  \small
  \begin{tabular}{lrrrr}
    \toprule
    \textbf{Taxonomy} & \textbf{F1} & \textbf{CD-L2} & \textbf{Final F1} & \textbf{Final CD-L2} \\
    \midrule
    Airplane & 0.323 & 0.033 & 0.326 & 0.032 \\
    Car      & 0.126 & 0.048 & 0.127 & 0.047 \\
    Chair    & 0.130 & 0.043 & 0.131 & 0.043 \\
    Lamp     & 0.288 & 0.035 & 0.288 & 0.035 \\
    Table    & 0.143 & 0.041 & 0.143 & 0.041 \\
    \bottomrule
  \end{tabular}
\end{table}

\begin{figure*}[t]
\centering
\includegraphics[width=0.85\textwidth]{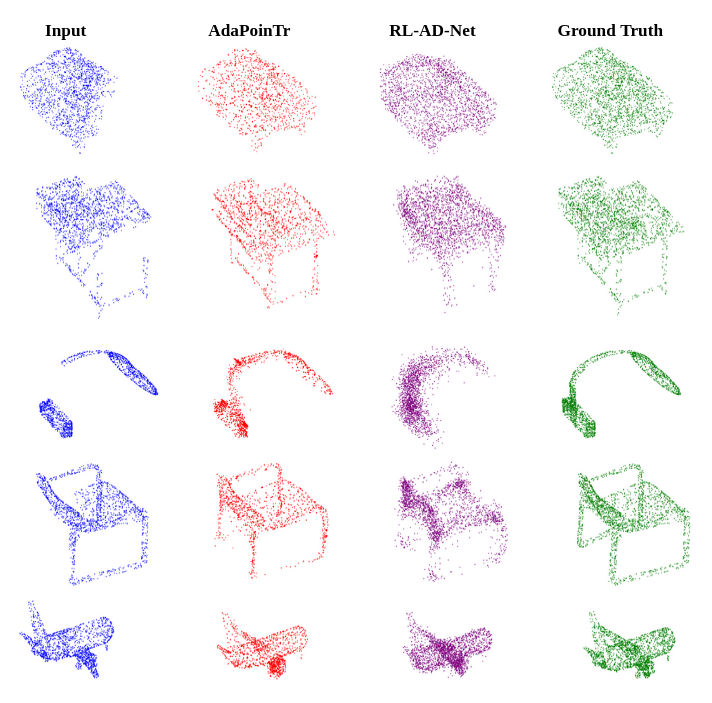}
\caption{\textbf{Qualitative failure cases under multi-category RL training.}  
Examples from \emph{Car},\emph{table},\emph{Lamp},\emph{Chair}, and \emph{Airplane} categories where the AdaPoinTr baseline (second column) produces visually cleaner and more plausible completions compared to the RL-AD-Net outputs (third column). This highlights the limitations of a single shared policy: without category specialization, the refinement tends to either over-smooth or distort valid structures, sometimes degrading the initial backbone completions rather than improving them. These observations align with the quantitative results in Table~\ref{tab:supp_multicat_results}, where only negligible gains are observed under multi-class RL training.}
\label{fig:supp_multicat_samples}
\end{figure*}

As shown in Table~\ref{tab:supp_multicat_results}, the per-sample selection module rarely favors the refined output under multi-category RL training, leading to nearly identical scores to the AdaPoinTr baseline. While small differences appear in the final metrics, these shifts are negligible relative to the overall dataset size and reflect the fact that only a handful of instances (3 to 6 per category, except for lamps) were selected as refined. We attribute this outcome to two main factors: (i) the heterogeneity of cross-category geometric priors, which introduces conflicts in the shared latent action space and prevents stable convergence; and (ii) the unsupervised nature of refinement, which complicates instance-level credit assignment without category-specific specialization. This result is consistent with the main paper, where \emph{category-wise} AE+RL training delivers measurable improvements, while a single multi-class RL head underfits the diverse shape manifold. Qualitative results (Figure~\ref{fig:supp_multicat_samples}) further show that the multi-class policy tends to propose conservative edits that are filtered out by the selector, reinforcing the necessity of category-specific refinement strategies.

\section{Miscellaneous: Point-NN for Geometry-Aware Selection}
\label{sec:supp_pointnn}

During deployment, ground truth point clouds are not available, so refinement must be evaluated without direct supervision. To this end, we integrate Point-NN, a non-parametric encoder that provides a geometry-aware quality score for completed shapes.

Point-NN extracts global descriptors from a point cloud by combining furthest point sampling (FPS), local neighborhood grouping via $k$-NN, positional encodings, and hierarchical pooling. This yields embeddings that capture both fine-grained local structure and overall shape coherence. Given two completions the AdaPoinTr baseline $P_{\text{base}}$ and the RL-refined output $P_{\text{ref}}$ Point-NN assigns normalized quality scores $q_{\text{base}}$ and $q_{\text{ref}}$. The final output is then selected as:
\[
P_{\text{out}} = 
\begin{cases} 
P_{\text{ref}}, & q_{\text{ref}} > q_{\text{base}}, \\ 
P_{\text{base}}, & \text{otherwise.}
\end{cases}
\]

This selection mechanism ensures that RL refinement is never detrimental: if the refinement introduces distortions, the system defaults to the backbone output; if it enhances structural fidelity, the refined version is chosen. When ground truth $P_{\text{gt}}$ is available, the selection further considers Chamfer Distance alongside Point-NN scores, combining geometric consistency and reconstruction accuracy. This dual criterion strengthens validation while guaranteeing robustness in deployment scenarios where ground truth is absent.

\clearpage
{
    \small
    \bibliographystyle{abbrvnat}  
    \bibliography{main}
}

\end{document}

%% file: preamble.tex
\usepackage[dvipsnames]{xcolor}
